# A PubMedBERT-based Classifier with Data Augmentation Strategy for Detecting Medication Mentions in Tweets

Qing Han, Shubo Tian, Jinfeng Zhang
Department of Statistics, Florida State University, Tallahassee, Florida, United States

*Abstract*—As a major social media platform, Twitter publishes a large number of user-generated text (tweets) on a daily basis. Mining such data can be used to address important social, public health, and emergency management issues that are infeasible through other means. An essential step in many text mining pipelines is named entity recognition (NER), which presents some special challenges for tweet data. Among them are nonstandard expressions, extreme imbalanced classes, and lack of context information, etc. The track 3 of BioCreative challenge VII (BC7) was organized to evaluate methods for detecting medication mentions in tweets. In this paper, we report our work on BC7 track 3, where we explored a *PubMedBERT*-based classifier trained with a combination of multiple data augmentation approaches. Our method achieved an F1 score of 0.762, which is substantially higher than the mean of all submissions (0.696).

## I. Introduction

With the global increase in social media usage, vast amount of information is generated every day. As a major social media communication platform, Twitter has 1.3 billion accounts and 330 million monthly active users posting 500 million tweets per day (3). Twitter users can generate their original tweets, "retweet" a tweet to show their reaction or comments, click the "like" button, tag another user's name, or respond to the author of the tweet (4). It records user-level prompt information and covers a wide range of different topic areas. As a representative application area, this global data source contains rich patient-generated information which can provide valuable insights for public health studies (7).

Although the importance of utilizing information from social media platforms like Twitter has gained more and more attention, constructing effective automatic named entity recognition and information extraction systems remains challenging. A common characteristic of tweet data is the nonstandard or semi-standard use of language by individual users. People pay less attention to spellings and precise linguistic development of a sentence in day-to-day discussions (1). This substantially increases various kinds of ambiguities, including lexical, syntactic, and semantic ambiguities (2). For example, arbitrary abbreviations with multiple potential meanings and the noise brought by non-standard inputs (special symbols, emojis and frequent typos etc.).

Another major challenge is identifying the relevant tweets from a large number of other irrelevant ones. This often leads to highly imbalanced corpora where the tweets of interest could account for less than 1% of all tweets in the corpora. Moreover, the length for each tweet is restricted to only 280 characters (140 characters prior to October 2018). This results in a limited context information which is essential for recognizing entities of interest.

With these challenges, the performance of many deep learning models that work well for most other NLP problems drop dramatically (11)(12)(7).

In this work, we experimented and fine-tuned several commonly used pre-trained transformer models including BERT and its variants. To cope with the highly unbalanced positive-negative case ratios, we used data augmentation to increase the proportion of positive cases and maximize the amount of information that could be extracted from limited labeled data (5)(6).

Our final model is a *PubMedBERT*-based classifier trained with a combination of multiple data augmentation approaches. Our method achieved very satisfactory performance with an F1 score of 0.762 from the best submission, which compared favorably to the mean of all submissions with an F1 score of 0.696.

## II. Data Description

The dataset consists of all tweets posted by 212 Twitter users during and after their pregnancy. This data represents the natural and highly imbalanced distribution of drug mentions on Twitter, with only approximately 0.2% of the tweets mentioning a medication.

Training data (BioCreative_TrainTask3.0 & 3.1) contains about 89,000 tweets with 218 tweets mentioning at least one drug. Validation data (BioCreative_ValTask3) contains 39,000 tweets with 93 tweets mentioning at least one drug. Additional data set comes from the "SMM4H'18 shared tasks" held in 2018 and it is a balanced dataset which contains about 10,000 tweets. And the final test datasets contains about 54,000 tweets with the similar positive vs. negative ratio as the training and validation datasets.

Please see the detail datasets composition in Table 1.

TABLE 1. DATASETS STATISTICS

| Dataset | Count of Unique Tweets | Positive | Negative | Count of Tweets containing more than 1 medications | Percentage of Positive Tweets |
|---|---|---|---|---|---|
| BioCreative_TrainTask3.0 | 49992 | 115 | 49877 | 7 | 0.23% |
| BioCreative_TrainTask3.1 | 38996 | 103 | 38893 | 8 | 0.26% |
| BioCreative_ValTask3 | 38137 | 93 | 38044 | 11 | 0.24% |
| SMM4H18_Train | 9622 | 4975 | 4647 | 677 | 51.7% |

## III. METHODS

### A. Preprocessing

Tweets consist of a large amount of nonstandard or semi-standard user-input sentences and characters including various special symbols and emojis. Cleaning special tokens and characters is necessary to remove noise and prepare the data for downstream model building. All tweets were pre-processed as following:

(1) Deleted all non-English characters, including emojis and special symbols that cannot be decoded by ASCII.

(2) Removed single characters if they are in the list of "#$%&@+*^`|~".

Please note that the position information of each character is recorded before doing any cleaning or removal.

(3) Lastly, we tokenized the sentences using the BERT standard tokenizer.

### B. Baseline Model Selection

To predict whether a tweet contains a medication or dietary supplement mention, we decided to use the state-of-the-art deep learning NLP framework, *BERT*-based classifiers (8).

Since its initial publication, the original BERT, has been extended with a diverse set of pre-trained models trained using various domain corpora. In biology and bioinformatics area, well-known pre-trained models include *BioBERT* (9), *PubMedBERT (abstract only)*, and *PubMedBERT (full text)* (10). We first compared the prediction results using all of these pretrained models with BioCreative_TrainTask3.0 dataset as the training set, BioCreative_ValTask3 dataset as the validation set, and BioCreative_TrainTask3.1 dataset as the test set (Table 2). With *PubMedBERT (full text)* achieving the highest F1 score, we decided to use *PubMedBERT (full text)* as our base model, which was further improved with data augmentation approaches.

TABLE 2. PRE-TRAINED MODELS COMPARISON AND SELECTION

| | Precision | Recall | F1 score |
|---|---|---|---|
| **BioBERT** | 0.7273 | 0.5045 | 0.5957 |
| **PubMedBERT (abs)** | 0.7794 | 0.4775 | 0.5922 |
| **PubMedBERT (full)** | 0.8235 | 0.5045 | 0.6257 |

Table 2. PubMedBERT perform better than BioBERT in terms of F1 score. Specifically, PubMedBERT version that is pre-trained on both abstracts and main body (full text) performed the best.
(Note: Training set: BioCreative_TrainTask3.0; Validation set: BioCreative_ValTask3; Test set: BioCreative_TrainTask3.1)

To increase the amount of training data, we also added additional data from SMM4H'18 into the training set, which significantly improved the performance (Table 3).

TABLE 3. PERFORMANCE CROSS-CHECK BY CHANGING TRAINING SET

| Training/Validation Sets | Test Set | Precision | Recall | F1 score |
|---|---|---|---|---|
| TrainTask3.0/ValTask3 | TrainTask3.1 | 0.8235 | 0.5045 | 0.6257 |
| TrainTask3.0&ValTask3&SMM4H18* | TrainTask3.1 | 0.7015 | 0.8468 | 0.7673 |
| TrainTask3.0&3.1* | ValTask3 | 0.7976 | 0.6381 | 0.7090 |
| TrainTask3.0&3.1&SMM4H18* | ValTask3 | 0.7165 | 0.8667 | 0.7845 |

*The splitting ratio of training and validation sets is 8:2.

### C. Data Augmentation strategy

To address the challenge with very limited number of tweets that contain at least one drug entity in the training data, we considered three main data augmentation strategies:

(1) Augment true cases by replacing each original true entity with a randomly chosen medication mention from the pool. The medication mention pool could be generated from either BioCreative_TrainTask3.0 & 3.1 two datasets, or BioCreative_TrainTask3.0 & 3.1 & SMM4H18 three datasets.

(2) Augment true cases by replacing each original true entity with a random string. The string contains 3 to 10 characters randomly selected from a-z and A-Z.

(3) Augment true cases by dropping a randomly selected word which is not or not belong to a true entity.

Each of the strategy has its own advantages in terms of adding additional useful information for the model training. The first strategy is relatively safe as it uses only the true entities, but it may not "squeeze out" enough information as using the random strings. The second strategy is more aggressive by forcing the model to learn more about the context information by introducing new names the pre-trained models have never seen before. But it may have introduced some unwanted bias. For example, the model may be trained to recognize unrelated random-string-like terms, such as serial numbers and abbreviations, as positive mentions. The third strategy introduces more diversity into the context itself although it may bring in more grammar mistakes or missing words. However, these are also the characteristics of tweets – irregular structure, nonstandard grammar, frequent missing words, and typos, etc.

For each of the strategies, there are options of augmenting all or partial data; and the number of rounds the data would be augmented based on positive cases. And combination of different strategies could also bring compound effects into the model training. Results for several exemplary experiments with different data augmentation strategies are showing in Table 4.

*D. Getting Additional Data*

In addition to augmentation of the original data sets, we also tried to add more training data by using certain "search terms" to scrape more tweets. The "search terms" used includes three parts: First, true entity pool mentioned in Strategy (1); Secondly, commonly used Over-the-Counter (OTC) drugs and supplements, which are manually selected from (13) and (14); Thirdly, commonly used drugs and supplements for pregnancy, which are manually selected from websites (15) and (16). In total, we scraped 13,844 valid tweets by 220 entity terms using "snscrape" package (17) in Python.

After getting the tweets, we did the same preprocessing procedure, and then added them into our training/validation sets. From the results in Table 5, we can see that the new tweets brought a big jump in Recall but lead to a lower Precision. This indicates that new data could significantly help in recognition of more entities so that false negative cases could be largely reduced with the cost of more false positive cases.

TABLE 4. DATA AUGMENTATION TRAINING EXAMPLES

| Strategy | Datasets for Augmentation | Precision | Recall | F1 score |
|---|---|---|---|---|
| (1)* x1 | TrainTask3.0 & 3.1 | 0.7787 | 0.9048 | 0.8370 |
| (1)* x3 | TrainTask3.0 & 3.1 | 0.9524 | 0.7407 | 0.8333 |
| (2) x1 | TrainTask3.0 & 3.1 | 0.7500 | 0.8571 | 0.8000 |
| (2) x1 | TrainTask3.0 & 3.1 & SMM4H18 | 0.7097 | 0.8381 | 0.7686 |
| (3) x1 | TrainTask3.0 & 3.1 | 0.7280 | 0.8667 | 0.7913 |
| (1)** x10 + (3) x1 | TrainTask3.0 & 3.1 | 0.7385 | 0.9143 | 0.8170 |

Table 4. The model was trained/validated with data including original sets of TrainTask3.0 & 3.1 & SMM4H18, and plus the augmentation set which is generated based on different datasets via different combination of data augment strategies.

*used the true entity pool that contains mentions from TrainTask3.0 & 3.1 datasets.
**used the true entity pool that contains mentions from TrainTask3.0 & 3.1 & SMM4H18 datasets.

TABLE 5. MODEL TRAINING EXAMPLES WITH ADDITIONAL NEW DATA

| Strategy | Training + Validation Sets | Precision | Recall | F1 score |
|---|---|---|---|---|
| NA | TrainTask3.0 & 3.1 | 0.7976 | 0.6381 | 0.7090 |
| NA | TrainTask3.0 & 3.1 + SMM4H18 | 0.7165 | 0.8667 | 0.7845 |
| NA | TrainTask3.0 & 3.1 + SMM4H18 + new | 0.6575 | 0.9143 | 0.7649 |
| (1)* x1 | TrainTask3.0 & 3.1 + Augment** + SMM4H18 + new | 0.7092 | 0.9524 | 0.8130 |
| (1)* x3 | TrainTask3.0 & 3.1 + Augment** + SMM4H18 + new | 0.6732 | 0.9810 | 0.7984 |

*used the true entity pool that contains mentions from TrainTask3.0 & 3.1 datasets.
**data augmentation performed on TrainTask3.0 & 3.1 datasets.

IV. CONCLUSION AND FINAL SUBMISSIONS

Based on the performances obtained using the training and validation datasets, we selected our three final models, which are listed in Table 6.

The first classifier utilized the *PubMedBERT (full text)* pretrained model and fine-tuned with the BioCreative_TrainTask3.0, BioCreative_TrainTask 3.1 and their augmented data sets, plus SMM4H'18 data set. Data augmentation strategy is the first strategy, which is "Augment 1 time of true case by replacing original true entity with a randomly chosen medication mention" where the medication mention pool is generated from only BioCreative_TrainTask3.0 & 3.1 data sets.

The second classifier utilized the *PubMedBERT (full text)* pretrained model and fine-tuned with the BioCreative_TrainTask3.0, BioCreative_TrainTask 3.1 and their augmented data sets, plus SMM4H'18 data set. The augmented data sets are generated by two strategies: the first one is "augment 10 times by replacing each of the original true entity with a randomly chosen medication mention" where the medication mention pool is generated from

BioCreative_TrainTask3.0 & 3.1 and SMM4H'18 three datasets. The second one is "augment 1 time by dropping a randomly selected word which is not or not belong to a true entity".

The third classifier utilized the *PubMedBERT (full text)* pretrained model and fine-tuned with the BioCreative_TrainTask3.0, BioCreative_TrainTask3.1 and their augmented datasets, plus the original SMM4H'18 data set. The augmented datasets were generated by two strategies: the first one is "augment 3 times by replacing each of the original true entity with a randomly chosen medication mention" where the medication mention pool is generated from only BioCreative_TrainTask3.0 & 3.1 data sets. The second one is "augment 1 time by dropping a randomly selected word which is not or not belong to a true entity".

The performances as shown in Table 7 indicate that the combined augmentation strategy could help the model learn more about the context information and further improve its entity recognition capability. Comparing performances of submission 1 and 3 shows that adding one more type of data augmentation strategy contributes to a higher Recall and less false negative. By randomly dropping out a word from the context of true entities, more variability is introduced into the training data and the performance could be further improved. However, the additional diversity could also be problematic if too much is introduced. For example, comparing submission 2 and 3, a lower precision indicates higher amount of false positives. Therefore, entities from SMM4H18 dataset increases the variability of entity recognitions and leads to less consistency with the test tweets which were also generated by the same 212 users.

TABLE 6. FINAL THREE SUBMISSIONS

| Submission | Training + Validation Sets* | Data Augmentation Rounds |
|---|---|---|
| 1 | TrainTask3.0 & 3.1 & Val3 + Augment by (1)** + SMM4H18 | (1) x1 |
| 2 | TrainTask3.0 & 3.1 & Val3 + Augment by (1)*** + TrainTask3.0 & 3.1 & Val3 + Augment by (3) + SMM4H18 | (1) x10 + (3) x1 |
| 3 | TrainTask3.0 & 3.1 & Val3 + Augment by (1)** + TrainTask3.0 & 3.1 & Val3 + Augment by (3) + SMM4H18 | (1) x3 + (3) x1 |

*The splitting ratio of training and validation sets is 8:2.
**used the true entity pool that contains mentions from TrainTask3.0&3.1&Val3 datasets.
***used the true entity pool that contains mentions from TrainTask3.0&3.1&Val3&SMM4H18 datasets.

TABLE 7. PERFORMANCES OF SUBMISSIONS

| Submission | Overlapping | | | Strict | | |
|---|---|---|---|---|---|---|
| | F1 score | Precision | Recall | F1 score | Precision | Recall |
| 1 | 0.764 | 0.747 | 0.782 | 0.738 | 0.721 | 0.755 |
| 2 | 0.763 | 0.712 | 0.823 | 0.732 | 0.682 | 0.789 |
| 3 | 0.794 | 0.744 | 0.85 | 0.762 | 0.714 | 0.816 |
| All Participants (mean ± std) | 0.749±0.0596 | 0.811 | 0.709 | 0.696±0.072 | 0.754 | 0.658 |